\documentclass[sigconf]{acmart}

\usepackage{algorithm}
\usepackage{algpseudocode}

\usepackage{comment}
\usepackage{soul}

\newcommand{\ignore}[1]{}
\newcommand{\subchange}[1]{{\color{black} #1}}

\usepackage{ulem}

\AtBeginDocument{%
  }

\copyrightyear{2024}
\acmYear{2024}
\setcopyright{rightsretained}
\acmConference[GeoAnomalies'24]{1st ACM SIGSPATIAL International Workshop on Geospatial Anomaly Detection}{October 29, 2024}{Atlanta, GA, USA}
\acmBooktitle{1st ACM SIGSPATIAL International Workshop on Geospatial Anomaly Detection (GeoAnomalies'24), October 29, 2024, Atlanta, GA, USA}
\acmDOI{10.1145/3681765.3698452}
\acmISBN{979-8-4007-1144-2/24/10}

\begin{document}

\title{ReeFRAME: Reeb Graph based Trajectory Analysis Framework to Capture Top-Down and Bottom-Up Patterns of Life}

\author{Chandrakanth Gudavalli}
\authornote{Equal Contributors.}
\affiliation{%
  \institution{ECE Department}
  \institution{University of California}
  \city{Santa Barbara}
  \country{USA}
}
\email{chandrakanth@ucsb.edu}

\author{Bowen Zhang}
\authornotemark[1]
\affiliation{%
  \institution{ECE Department}
  \institution{University of California}
  \city{Santa Barbara}
  \country{USA}
}
\email{bowen68@ucsb.edu}

\author{Connor Levenson}
\affiliation{%
  \institution{ECE Department}
  \institution{University of California}
  \city{Santa Barbara}
  \country{USA}
}

\email{clevenson@ucsb.edu}

\author{Kin Gwn Lore}
\affiliation{%
  \institution{RTX Technology Research Center}
  \state{Connecticut}
  \country{USA}
}
  \email{kin.lore@rtx.com}

\author{B. S. Manjunath}
\affiliation{%
  \institution{ECE Department}
  \institution{University of California}
  \city{Santa Barbara}
  \country{USA}
}
\email{manj@ucsb.edu}




\renewcommand{\shortauthors}{Gudavalli et al.}

\begin{abstract}
In this paper, we present ReeFRAME, a scalable Reeb graph-based framework designed to analyze vast volumes of GPS-enabled human trajectory data generated at 1Hz frequency. ReeFRAME models Patterns-of-life (PoL) at both the population and individual levels, utilizing Multi-Agent Reeb Graphs (MARGs) for population-level patterns and Temporal Reeb Graphs (TERGs) for individual trajectories. The framework's linear algorithmic complexity relative to the number of time points ensures scalability for anomaly detection. We validate ReeFRAME on six large-scale anomaly detection datasets, simulating real-time patterns with up to 500,000 agents over two months.
\end{abstract}




\received{8 September 2024}
\received[revised]{1 October 2024}
\received[accepted]{5 October 2024}

\maketitle

\section{Introduction}
\label{sec:intro}
\begin{figure}[t]
    \centering
    \includegraphics[width=1.0\columnwidth]{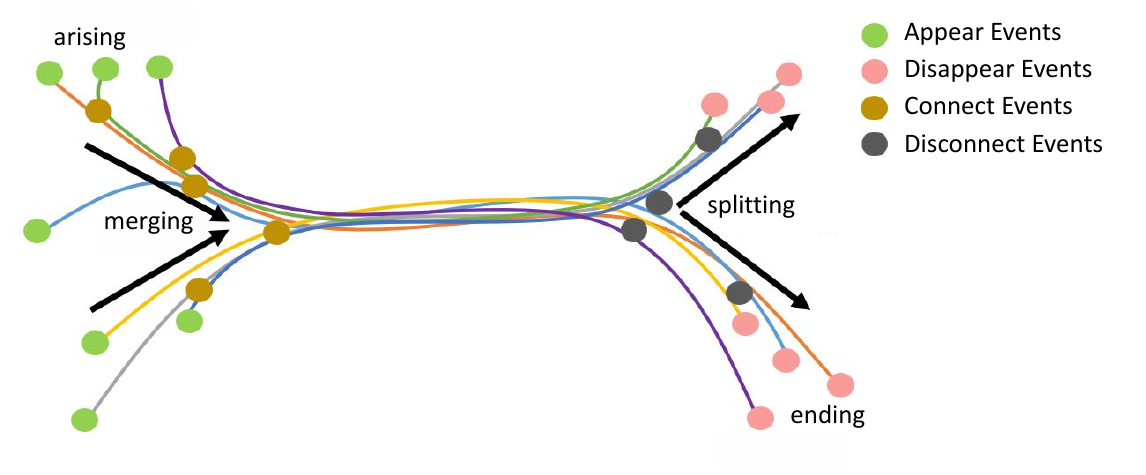}
    \caption{Set of seven one-day trajectories of an agent, contributing to appear/disappear/connect/disconnect events. Events will be later used to construct Reeb graph~\cite{shailja2021computational} for the agent.}
    \label{fig:reeb_intro}
    \vspace{-0.5cm}
\end{figure}
Trajectories encapsulate the dynamic nature of movement across various domains, ranging from individual human activities to vehicular traffic and migratory patterns of animals. As such, they are ubiquitous and integral to understanding complex systems in urban planning, ecology, and social sciences. However, modeling these trajectories to extract meaningful patterns and detect anomalies is a significant challenge that requires sophisticated analytical tools and methods. This challenge is compounded by the magnitude and complexity of trajectory data, which often involves irregular sampling rates, noise, and multidimensional attributes~\cite{pentland2009inferring}.

The analysis of trajectories involves identifying key events, such as points where trajectories converge (connect events) or diverge (disconnect events), as shown in Figure~\ref{fig:reeb_intro}. These events are crucial to understanding group behavior, traffic flow interruptions, and other critical dynamics within the studied systems. The ability to detect and analyze these events can provide insights into routine and anomalous behaviors, aiding in everything from traffic management to predictive policing~\cite{zheng2008learning}. However, traditional methods of trajectory analysis often struggle with the volume and complexity of data, failing to efficiently process and interpret large-scale datasets.

Reeb graphs offer a robust solution to these challenges. Originating from topology, Reeb graphs abstract the continuity and connectivity of data, making them particularly suited for analyzing complex trajectory data~\cite{shinagawa1991surface}. By focusing on the essential structure of data and filtering out noise and irrelevant details, Reeb graphs simplify the representation of trajectories, capturing critical transitions and changes in the connectivity patterns.



\subchange{
Human trajectories are shaped by a combination of top-down and bottom-up societal patterns. \textit{Top-down patterns} arise from external structures imposed by institutions or societal norms, such as city infrastructure, traffic regulations, or work schedules, which guide and constrain movement at the population level. These patterns manifest in organized, predictable behaviors like daily commutes or visits to public spaces constrained by operating hours. In contrast, \textit{bottom-up patterns} emerge from individual-level decisions driven by personal needs and spontaneous choices. These behaviors are more dynamic and less predictable, shaped by factors like individual preferences, spontaneous travel, or personal routines. Our framework, ReeFRAME, captures both types of patterns by leveraging \textit{Multi-Agent Reeb Graphs (MARGs)} to model top-down population-level structures and \textit{Temporal Reeb Graphs (TERGs)} to analyze bottom-up, agent-specific behaviors. By combining these approaches, ReeFRAME provides a comprehensive tool for detecting anomalies in human trajectories, identifying deviations from both societal norms and individual consistency.
}

Traditional Reeb graphs, typically applied in two-dimensional spaces, are useful for understanding the spatial continuity and connectivity of individual trajectories. However, they have limitations: they do not encode temporal information, are not well-suited for trajectories with loops, and have a high computational complexity of $O(n^2)$. To address these issues, time-based Reeb graphs \cite{zhang2024reespot} extend the analysis to include temporal dimensions by constructing Reeb events at different timestamps, with a computational complexity of $O(n)$. This approach allows for the examination of how trajectories evolve over time and supports the incorporation of multiple features at each node of the Reeb graph. When dealing with complex systems involving multiple agents, such as urban traffic, Multi-Agent Reeb Graphs (MARG) offer a scalable and interpretable framework that captures population-level patterns while preserving the ability to analyze individual trajectories. By integrating these various forms of Reeb graphs, our approach provides a comprehensive toolset for analyzing trajectories at both micro and macro levels.

\noindent\textbf{Main Contributions:}
\begin{itemize}
    \item We propose a generalizable framework that leverages Reeb graphs to model both top-down and bottom-up patterns of normalcy in trajectory data.
    \item We introduce Multi-Agent Reeb Graphs (MARG), a scalable methodology designed to represent population-level patterns within an interpretable Reeb graph structure.
    \item We outline techniques to enhance the descriptiveness of agent-level Temporal Reeb Graphs (TERG), enabling them to capture bottom-up patterns of normalcy.
    \item We validate our approach using a large-scale, simulated GPS trajectory dataset involving up to 500,000 agents in a single city over a two-month period, demonstrating its effectiveness in capturing and analyzing complex patterns at scale.
\end{itemize}

\section{Related Work}
\label{sec:rel_work}
The analysis and modeling of human mobility patterns have garnered considerable interest due to the widespread adoption of GPS-enabled devices. Recent advances in location-based services have accelerated consumer incorporation of smart devices (e.g., smartphones and wearables) into activities and routines in daily lives. As GPS positional data provides the foundation to reveal various behavioral patterns present in a person, it has been frequently adopted to model the movements of a suspect in relation to normal population behavior.
Traditional approaches often rely on geometric features and statistical techniques to model trajectory data. Studies such as those by ~\cite{zheng2008learning} and ~\cite{zhang2019mining} have utilized statistical methods to extract patterns like mean velocity and periodic behaviors from trajectory datasets. However, these methods struggle with the complexity and dynamic nature of high-dimensional mobility data, as they cannot adequately capture the intricate spatial relationships and temporal regularities inherent in human movements.

Advancements in machine learning, particularly deep learning, have introduced more sophisticated models for trajectory analysis. For instance, neural networks like LSTMs and attention-based models have been explored for next-location prediction and anomaly detection in movement patterns, providing enhanced accuracy but at the cost of interpretability~\cite{luca2021survey, zeng2019next}. Despite their efficacy, the black-box nature of these models limits their practical application, especially in scenarios requiring transparent decision-making processes. 
Additionally, anomaly detection often involves heavily skewed data distributions, where normal behavior constitutes the majority of the data and anomalies are rare events (0.1\% anomaly to normal ratio in our datasets). This imbalance poses significant challenges for model training and evaluation. The small anomaly sample size makes it difficult for a model to learn a generalized representation of what constitutes an anomaly. This diversity among anomalies can lead to a situation where the model fails to detect novel or previously unseen anomalies~\cite{pang2021deep}.

Graph-based methods have emerged as powerful tools for representing complex spatial interactions and temporal transitions, as illustrated by~\cite{guo2010graph, qi2015efficient}. 
These methods encapsulate relationships among trajectories in a more intuitive and interpretable manner, allowing for the exploration of both micro and macro mobility behaviors. 

However, scaling 
these methods to handle large volumes of high-dimensional data efficiently presents significant challenges. Deep learning models require substantial computational resources and are prone to overfitting and lengthy training times due to their complex, multi-layered architectures~\cite{wang2017knowledge}. Similarly, graph-based methods struggle with managing large graphs, dynamically updating them, and optimizing graph partitioning to balance computational loads effectively~\cite{kipf2016semi}. Addressing these scalability issues involves strategies such as model simplification, use of distributed and parallel computing, and incremental learning to update models without full retraining. Additionally, approximation algorithms can reduce computational demands, while leveraging AWS~\cite{aws_parallel_computing} cloud parallel computing and specialized hardware like CPUs or GPUs can significantly enhance processing efficiency and scalability. Combining these approaches enables scaling of these technologies to meet the demands of large-scale applications, maintaining performance without sacrificing speed or accuracy.


This paper explores the use of Reeb graphs for analyzing mobility \subchange{patterns}. Reeb graphs have been employed to model and analyze the structure of white matter pathways in the brain, offering insights into neural connectivity and brain architecture~\cite{shailja2023reebundle, shailja2023retrace}. Similarly, in computational geometry, Reeb graphs facilitate the analysis of shape and form, providing tools for shape segmentation and recognition, which are essential in fields such as computer-aided design and manufacturing.

ReeSPOT~\cite{zhang2024reespot} builds upon this foundation by employing Reeb graphs, which abstract and simplify trajectory data into a topological structure that captures significant spatial and temporal deviations. This approach not only enhances the interpretability of the data but also scales efficiently with the volume of data points, addressing a critical gap in traditional and machine learning-based methods. By clustering common behavior patterns using Reeb graphs, ReeSPOT extends the capabilities of graph-based analysis to effectively model and identify deviations from normalcy in human trajectories, setting a new direction for research in the domain of agent behavioral analysis and anomaly detection. However, ReeSPOT~\cite{zhang2024reespot} is not designed to capture population level patterns, which can be very useful to model human mobility signatures.

\begin{figure*}[t]
    \centering
    \includegraphics[width=2.1\columnwidth]{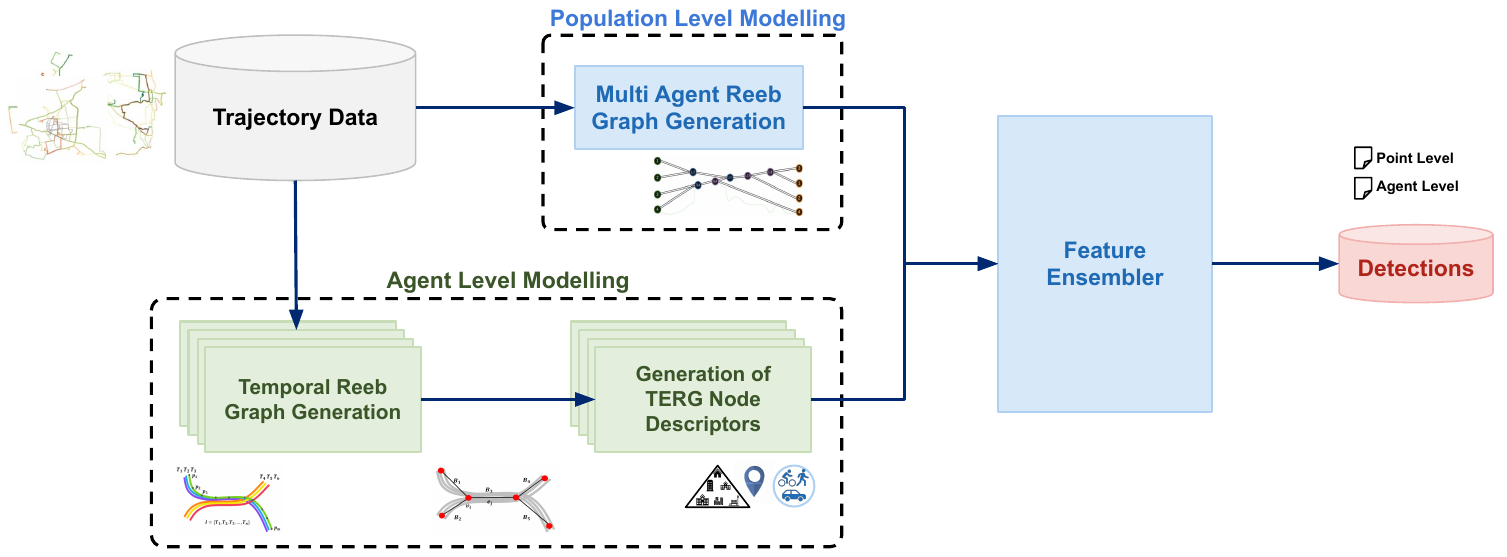}
    \caption{Architectural overview of the proposed anomaly detection framework, ReeFRAME. The pipeline splits into two key modeling processes: \textit{Agent Level Modeling}, where Temporal Reeb Graphs (TERGs) are generated for individual agents, and \textit{Population Level Modeling}, where a Multi-Agent Reeb Graph (MARG) is constructed to model broader population patterns. The TERG and MARG features are combined in the Feature Ensembler, which integrates and analyzes the data to produce Detections, identifying agents that deviate from normal behavior.}
    \label{fig:reeb_block}
\end{figure*}

To tackle the challenges of scalability while effectively capturing both agent-specific and population-level patterns, we introduce ReeFRAME. This framework addresses the shortcomings of existing methods, providing a more powerful approach for modeling and analyzing human mobility patterns at both micro and macro levels.

\section{Methodology}
\label{sec:methodology}

In this section, we provide detailed descriptions of the algorithms executed by each module within our framework. A visual representation of the architectural overview of ReeFRAME is provided in Figure~\ref{fig:reeb_block}. Section~\ref{ssec:tergs} and~\ref{ssec:time_reebs_desc} detail the algorithms used for agent-level modeling, while Section~\ref{sec:incrementalreeb} addresses the algorithms employed for population-level modeling. The methods for integrating agent-level and population-level models are discussed in Section~\ref{ssec:fusion}.


\subsection{Temporal Reeb Graph Construction (TERG)}
\label{ssec:tergs}

A Reeb graph is a mathematical structure used to analyze the topology of a manifold. It was first introduced as a means to study the evolution of level sets of a real-valued function defined on a manifold~\cite{shinagawa1991surface}. The nodes of a Reeb graph represent critical points where the topology of the level sets changes, such as merges, splits, or holes. The edges of the graph connect these nodes, indicating continuous changes between critical points. Reeb graph on the spatial data was first used in \cite{buchin2013trajectory} to represent the merging and splitting structures. 

The concept of Reeb graphs applied to trajectory data is significant in modeling the patterns of daily human activities by capturing and analyzing the regular, predictable patterns of movement that characterize daily human behavior, thus, enabling the potential to detect deviated anomalous trajectories from these patterns. The discussed work ReeSPOT \cite{zhang2024reespot} leverages the structure of Reeb graphs to interpret and analyze simulated human movements via GPS trajectories.
GPS trajectories are defined as sequences of time-stamped GPS coordinates representing individual movements:
\[
T = \{t_0: p_0, t_1: p_1, \ldots, t_m: p_m\}
\]
where \( t_i \) is a timestamp and \( p_i = (lat_i, long_i) \) denotes the geographical position at time \( t_i \).

The formulation of Temporal Reeb Graph (TERG), \( R(V, E) \), involves mapping trajectories to a topological space where nodes \( V \) represent significant events or critical points, and edges \( E \) denote the continuity of movement between these nodes. 

The constructions of Reeb graphs are described in Algorithm 1 and 2 in the ReeSPOT paper~\cite{zhang2024reespot}. Reeb graphs are constructed by first finding connect and disconnect events based on a distance threshold \(\epsilon\). 
The Euclidean distance, d, between 2 GPS points is as follows:
\[ d(p_i, p_j) = \sqrt{(lat_i - lat_j)^2 + (long_i - long_j)^2}. \]
For example, appear and disappear events represent the start and ending of trajectories. At a timestamp \(k\), \subchange{given two subtrajectories of an agent $T$ and $T'$,} if \( d(T[k], T'[k]) < \epsilon \), it is a connect event, and if \( d(T[k], T'[k]) \geq \epsilon \), we mark it as a disconnect event. One can use haversine distance instead of euclidean distance, depending upon the vastness of area of interest (AoI).
A Reeb graph is then constructed using the detected events by three steps:
\begin{itemize}
    \item \textbf{Event computation} 
    The first step involves computing events including appear, disappear, connect, and disconnect, where the algorithm determines potential events at each time point in the trajectories \( T \) and \( T' \). This computation takes \( O(m) \) time, where \( m \) represents the number of timestamp points in the trajectories.

    \item \textbf{Dynamic Graph Construction}
    The next step is to construct the dynamic graph \( G \) using the computed events. The dynamic graph is updated at each timestamp \( G_k \), where \( k \in \{1, 2, \dots, K\} \). Nodes in \( G_k \) represent daily trajectories, and edges represent \( \epsilon \)-connectivity between them, reflecting the proximity of trajectories at each time point.

    \item \textbf{Bundle Partition} 
    The connected components of \( G_k \) are grouped into bundles, denoted as \( B_i \). This grouping is achieved using graph traversal techniques to identify components that exhibit direct connectivity. The same bundle appearing at different continuous timestamps indicates that the trajectories remain unchanged, while unique bundles signify where events or changes are occurring.

    \item \textbf{Reeb Graph Construction} 
    Finally, the Reeb graph assembly constructs the final Reeb graph \( R(V, E) \) from the identified bundles. Each node in \( R(V, E) \) represents a unique bundle \( B_i \), and the edges between the nodes represent the transitions between these bundles. The assembly process highlights the primary pathways and transitions in the data, simplifying the complex trajectory data into a topologically meaningful and manageable structure.

\end{itemize}

\subsection{Making Temporal Reebs Describable}
\label{ssec:time_reebs_desc}

Now that we have a scalable way to capture an agent’s trajectory into a Reeb graph, we can create a clear summary of their movements. This summary helps us identify key moments when the agent’s daily sub-trajectories connect, cluster, or disconnect from each other. By focusing on these important events, we filter out most of the redundant information in the trajectory. To perform further analysis on these Reeb graphs, it's important to describe the nodes and edges effectively.

For trajectory analysis, we compute the following feature types:

\begin{enumerate}
    \item \textbf{Location-based Features:} These include the agent’s latitude and longitude, along with the timestamp at each GPS coordinate.
    \item \textbf{Kinematic Features:} These features include the agent’s velocity at a given node, the estimated mode of transportation, travel direction, the average stop duration at each location and radius of gyration (which is the diameter of both training and testing Reeb graphs). Here we define diameter as the maximum of haversine distances between all pair of nodes in the reeb graph.
    \item \textbf{Semantic Features:} These describe Points of Interest (POIs) around a Reeb node, such as home, work, or other significant locations.
\end{enumerate}

Each node within the agent-level Reeb graph is described using these features. This enabled us to train various machine learning models that can operate on TERGs.

\begin{figure*}[t]
    \centering
    \includegraphics[width=2.1\columnwidth]{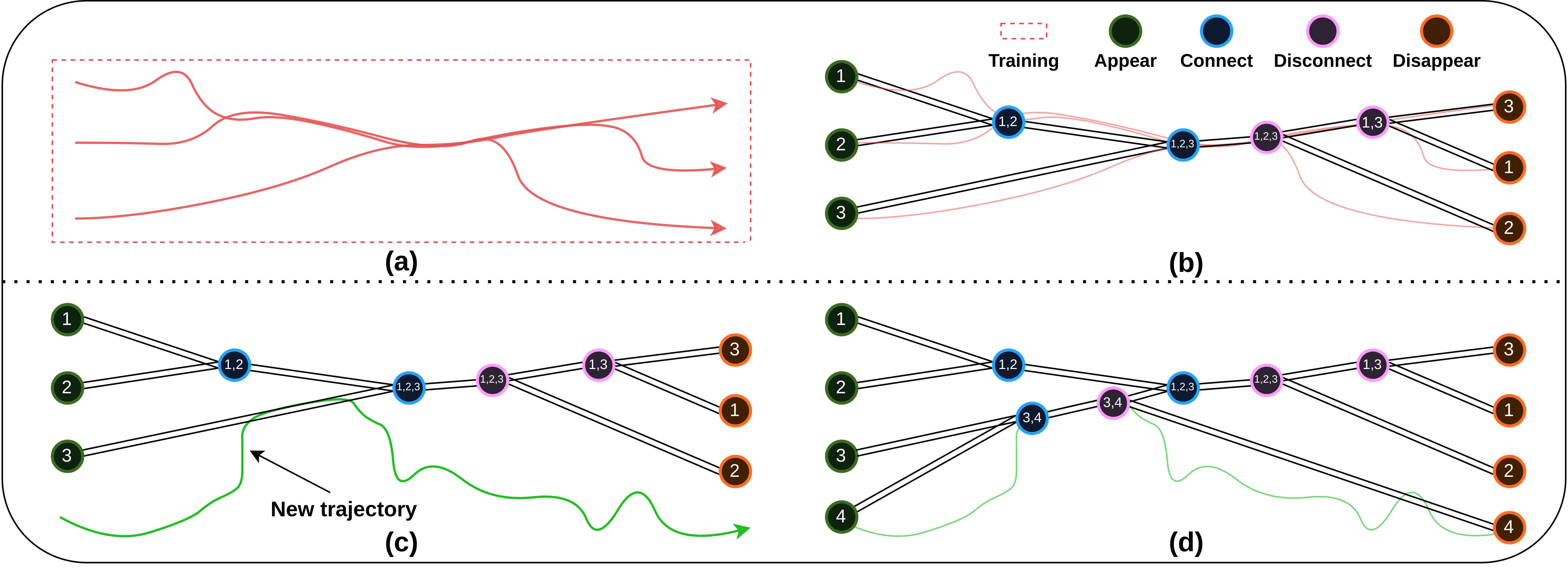}
    \caption{Demonstration of Incremental Reeb graph Construction using a toy example. (a) Sample set of three sub-trajectories; (b) Reeb graph constructed from the sub-trajectories; (c)~A new sub-trajectory that will be introduced into the Reeb graph; (d)~Updated Reeb graph accommodating the new sub-trajectory.}
    \label{fig:mag_reeb}
\end{figure*}

\subsection{Multi-Agent Reeb Graphs (MARGs)}
\label{sec:incrementalreeb}

While Temporal Reeb Graphs (TERGs) effectively capture individual-specific patterns, they do not account for population-level patterns, which are essential for modeling broader patterns of normalcy in a society. To address this, we propose the use of Multi-Agent Reeb Graphs (MARGs), which model the trajectories of an entire population within a single Reeb graph. The core idea is to construct MARGs using training trajectories and then compare the TERG generated for each agent's test trajectory against the pre-computed MARG.

However, the computation of MARGs can be time-consuming if we treat all agents as a list of sub-trajectories and compute a temporal Reeb graph for the entire corpus. 
\subchange{Given a population of $N$ agents, each with $k$ sub-trajectories, the time complexity for computing MARGs is $O(L \cdot N \cdot \log(N))$, where $L$ represents the length of each sub-trajectory.}

To mitigate this potential increase in computation time, we compute MARGs incrementally, as illustrated in Figure~\ref{fig:mag_reeb}. For example, if a Reeb graph has been computed for one agent over a 30-day period, we can use the Incremental Reeb Graph (IRG) construction methodology, as outlined in Algorithm~\ref{alg:IncrementalReebGraph}, to add a new one-day trajectory to the existing Reeb graph. We have extended this concept from a single agent to a population of agents to compute MARGs incrementally. Initially, we compute the temporal Reeb graph for $M$ agents. In our experiments, we set $M = 1000$, which allows us to compute the initial Reeb graph. Using this as a starting point, we then incrementally add the rest of the population’s sub-trajectories to the initial Reeb graph. The time complexity for adding a sub-trajectory to a Reeb graph is linear with respect to the length of the sub-trajectory ($L$) and the number of agents ($N$) in the Reeb graph, $O(L \cdot N)$.

Although we developed MARGs using the Incremental Reeb Graph (IRG) approach, IRGs can also be a valuable tool for updating an existing Reeb graph. For instance, if a Reeb graph has been constructed for an agent using 30 one-day sub-trajectories, the incremental Reeb approach described in this section can update the agent’s existing Reeb graph. Similarly, if a Reeb graph has been computed for 1000 agents, another agent can be added to the existing Reeb graph using the IRG algorithm. This approach avoids the need to recompute the Reeb graph for the entire population when adding just one more agent. This paper presents a proof-of-concept approach for developing IRGs, but more extensive large-scale experiments are needed to apply this idea to trajectories at scale in various contexts.

\begin{algorithm}
\caption{Incremental Reeb Graph Generation}
\label{alg:IncrementalReebGraph}
\begin{algorithmic}[1]
\State \textbf{Input:} Existing Reeb graph $G$, Sub-Trajectory $T$, Threshold $\epsilon$
\State \textbf{Output:} Updated Reeb graph $G'$
\State 
\Procedure{UpdateReeb}{$G, T, \epsilon$}
    \State Initialize $G' \gets G$
    \State Initialize Event List $E \gets []$
    \State Initialize $connected \gets \textbf{false}$
    \For{each point $p_i$ at time $t_i$ in $T$}
        \For{each edge $e$ in $G'$}
            \State Compute the closest point $p_e$ on edge $e$ at time $t_i$
            \State Calculate distance $d$ between $p_i$ and $p_e$
            \If{$d < \epsilon$ \textbf{and} $connected = \textbf{false}$}
                \State Add a connect event to $E$ 
                \State Set $connected \gets \textbf{true}$
            \ElsIf{$d \geq \epsilon$ \textbf{and} $connected = \textbf{true}$}
                \State Add a disconnect event to $E$
                \State Set $connected \gets \textbf{false}$
            \EndIf
        \EndFor
    \EndFor
    \For{each event $e$ in $E$}
        \State Update $G'$ based on the event $e$
    \EndFor
    \State \textbf{return} $G'$
\EndProcedure
\end{algorithmic}
\end{algorithm}

\subsection{Fusing TERGs with MARGs}
\label{ssec:fusion}
To effectively analyze the trajectories, we construct Temporal Reeb Graphs (TERG) for agent level and Multi Agent Reeb Graphs (MARG) for population level analysis. Let \( R_i = (V_i, E_i) \) represent the TERG for the \( i \)-th agent's training data, where nodes \( V_i \) capture significant events or changes in the agent's trajectory, and edges \( E_i \) represent continuous movement between these events. Similarly, the MARG \( R_G = (V_G, E_G) \) is constructed by aggregating the trajectories of all agents (as described in Section~\ref{sec:incrementalreeb}), with nodes \( V_G \) representing events common across multiple agents, and edges \( E_G \) denoting the flow of movement at the population level.

\subsubsection{Test Trajectory Analysis}

When a new test trajectory \( T_{test} \) is introduced, a corresponding test Reeb graph \( R_{test} = (V_{test}, E_{test}) \) is generated using the same methodology as for the agent-level Reeb graphs. This graph captures the key events and transitions within the test trajectory.

Each node \( v_{test} \in V_{test} \) in the test Reeb graph is compared separately with the agent-level Reeb graph \( R_i \) and the global-level Reeb graph \( R_G \), where \( i \) is the same agent as the test Reeb graph. A similarity score \( S_{i}(v_{test}) \) is computed for the agent-level comparison by evaluating the proximity and structural similarity between \( v_{test} \) and the closest node in \( R_i \). Similarly, for the global-level comparison, a score \( S_G(v_{test}) \) is calculated by comparing \( v_{test} \) with the closest node in \( R_G \).

\subsubsection{Combining Scores}

After obtaining the scores \( S_{i}(v_{test}) \) and \( S_G(v_{test}) \) for each node \( v_{test} \) in the test Reeb graph, these scores are combined to produce a final score \( S(v_{test}) \). The final score can be computed as a weighted sum of the individual scores:

\[
S(v_{test}) = \alpha \cdot S_{i}(v_{test}) + \beta \cdot S_G(v_{test})
\]

\noindent
where \( \alpha \) and \( \beta \) are weights that determine the relative importance of the agent-level and population-level comparisons for different scenarios. This combined score \( S(v_{test}) \) reflects both the individual and population-level consistency of the test trajectory, enabling robust anomaly detection. 
\subchange{The selection of \( \alpha \) and \( \beta \) is dataset specific, with their values determined empirically based on the characteristics of the datasets, as detailed in Section~\ref{ssec:feat_engg}.}

\begin{table*}[ht]
\begin{tabular}{|c|c|c|c|c|c|c|}
\hline
                                                                                                           & \textbf{D-1.1}                                               & \textbf{D-1.2}                                               & \textbf{D-1.3}                                               & \textbf{D-1.4}                                               & \textbf{D-2.1}                                               & \textbf{D-2.2}                                               \\ \hline
\textbf{Total number of agents}                                                                            & 200,000                                                      & 200,000                                                      & 200,000                                                      & 200,000                                                      & 500,000                                                      & 500,000                                                      \\ \hline
\textbf{Total number of points}                                                                            & \begin{tabular}[c]{@{}c@{}}1.03 \\ Trillion\end{tabular}     & \begin{tabular}[c]{@{}c@{}}1.03 \\ Trillion\end{tabular}     & \begin{tabular}[c]{@{}c@{}}1.03 \\ Trillion\end{tabular}     & \begin{tabular}[c]{@{}c@{}}1.03 \\ Trillion\end{tabular}     & \begin{tabular}[c]{@{}c@{}}864\\ Billion\end{tabular}        & \begin{tabular}[c]{@{}c@{}}864\\ Billion\end{tabular}        \\ \hline
\textbf{Sampling Frequency}                                                                                & 1 Hz                                                         & 1 Hz                                                         & 1 Hz                                                         & 1 Hz                                                         & 0.33 Hz                                                      & 0.33 Hz                                                      \\ \hline
\textbf{\begin{tabular}[c]{@{}c@{}}Number of \\ Anomaly Generation Teams\end{tabular}}                     & 9                                                            & 9                                                            & 9                                                            & 9                                                            & 4                                                            & 4                                                            \\ \hline
\textbf{\begin{tabular}[c]{@{}c@{}}Number of Anomalous agents \\ per Anomaly Generation Team\end{tabular}} & 30                                                           & 30                                                           & 30                                                           & 30                                                           & 100                                                          & 100                                                          \\ \hline
\textbf{\begin{tabular}[c]{@{}c@{}}Number of \\ anomalous agents\end{tabular}}                             & \begin{tabular}[c]{@{}c@{}}$\sim$270 \\ (30x9)\end{tabular}  & \begin{tabular}[c]{@{}c@{}}$\sim$270 \\ (30x9)\end{tabular}  & \begin{tabular}[c]{@{}c@{}}$\sim$270 \\ (30x9)\end{tabular}  & \begin{tabular}[c]{@{}c@{}}$\sim$270 \\ (30x9)\end{tabular}  & \begin{tabular}[c]{@{}c@{}}$\sim$400 \\ (100x4)\end{tabular} & \begin{tabular}[c]{@{}c@{}}$\sim$400 \\ (100x4)\end{tabular} \\ \hline
\textbf{\begin{tabular}[c]{@{}c@{}}Total number of \\ anomalous points\end{tabular}}                       & 1.01M                                                        & 0.95M                                                        & 0.8M                                                         & 0.98M                                                        & 1.88M                                                        & 1.93M                                                        \\ \hline
\textbf{\begin{tabular}[c]{@{}c@{}}Number of agents \\ in each Weak Label Group\end{tabular}}              & 300                                                          & 300                                                          & 300                                                          & 300                                                          & 800                                                          & 800                                                          \\ \hline
\textbf{Duration of Train Trajectories}                                                                    & \begin{tabular}[c]{@{}c@{}}1 month \\ (31 days)\end{tabular} & \begin{tabular}[c]{@{}c@{}}1 month \\ (28 days)\end{tabular} & \begin{tabular}[c]{@{}c@{}}1 month \\ (28 days)\end{tabular} & \begin{tabular}[c]{@{}c@{}}1 month \\ (28 days)\end{tabular} & \begin{tabular}[c]{@{}c@{}}1 month \\ (31 days)\end{tabular} & \begin{tabular}[c]{@{}c@{}}1 month \\ (28 days)\end{tabular} \\ \hline
\textbf{Duration of Test Trajectories}                                                                     & \begin{tabular}[c]{@{}c@{}}1 month \\ (31 days)\end{tabular} & \begin{tabular}[c]{@{}c@{}}1 month \\ (28 days)\end{tabular} & \begin{tabular}[c]{@{}c@{}}1 month \\ (28 days)\end{tabular} & \begin{tabular}[c]{@{}c@{}}1 month \\ (28 days)\end{tabular} & \begin{tabular}[c]{@{}c@{}}1 month \\ (31 days)\end{tabular} & \begin{tabular}[c]{@{}c@{}}1 month \\ (28 days)\end{tabular} \\ \hline
\textbf{\begin{tabular}[c]{@{}c@{}}Modes of \\ Transportation\end{tabular}}                                & Car                                                          & Car                                                          & Car                                                          & Car                                                          & \begin{tabular}[c]{@{}c@{}}Car/Bike/\\ Walk\end{tabular}     & \begin{tabular}[c]{@{}c@{}}Car/Bike/\\ Walk\end{tabular}     \\ \hline
\textbf{Size of Area of Interest}                                                                          & \begin{tabular}[c]{@{}c@{}}237 \\ sq km\end{tabular}         & \begin{tabular}[c]{@{}c@{}}15540 \\ sq km\end{tabular}       & \begin{tabular}[c]{@{}c@{}}1126 \\ sq km\end{tabular}        & \begin{tabular}[c]{@{}c@{}}19662 \\ sq km\end{tabular}       & \begin{tabular}[c]{@{}c@{}}281.6 \\ sq km\end{tabular}       & \begin{tabular}[c]{@{}c@{}}281.6 \\ sq km\end{tabular}       \\ \hline
\end{tabular}

\caption{Detailed description of the datasets that are used to test ReeFRAME for anomaly detection.} 
\label{tab:dataset_details}
\end{table*}

\section{Experiments}
\label{sec:experiments}

We applied our pipeline, ReeFRAME, to perform anomaly detection on four datasets of simulated human trajectories. In this section, we demonstrate the effectiveness of ReeFRAME in identifying anomalous agents at scale. Section~\ref{ssec:datasets} details the datasets used for the anomaly detection tasks. In Section~\ref{ssec:feat_engg}, we present both the quantitative and qualitative results across all four datasets, following a discussion on the feature engineering applied to the Reeb graphs. Finally, Section~\ref{ssec:aws} outlines the strategies employed to scale the anomaly detection inference and to prepare ReeFRAME as a cloud-ready framework.

\subsection{Datasets}
\label{ssec:datasets}

In this paper, we present the results of our anomaly detection method applied to four datasets containing trajectories of various individuals, referred to as \textit{agents}. We also describe how we simulated both normal and anomalous agent behaviors, along with the assumptions made during dataset generation. In Section~\ref{ssec:feat_engg}, we share the quantitative and qualitative performance of ReeFRAME on these datasets.

Each of the four datasets was specifically created to test anomaly detection. As shown in Table~\ref{tab:dataset_details}, the datasets simulate two consecutive months of activity in a hypothetical city. The first month is used as a training period, where all agents are assumed to behave normally, providing a baseline for what typical behavior looks like. During the second month, a small percentage of agents (< 0.01\% of the population) engage in unusual activities for varying periods. The normal trajectories for both the training and testing phases were generated by one team, while a separate set of teams were tasked with creating the anomalous trajectories.

The datasets were generated with sets of simulation parameters called 'trials' - each with different duration, sampling rate, modes of transportation, and areas of interest, with increasing complexity. Trials were carried out by multiple teams with their own technical approach to generate anomalies (to remain undetected by other teams) and to detect anomalies (generated by other teams). 
To generate these anomalies, the teams were provided with the full set of training data along with specific instructions, such as directing certain agents to deviate from their usual paths to enter and exit designated areas within specific time frames. Each team was assigned a different set of agents to alter. All agents generated in this manner were labeled as anomalous. The teams worked independently to ensure that the normal and anomalous data remained unbiased.

\begin{figure*}[!htbp]
    \centering
    \includegraphics[width=0.99\linewidth]{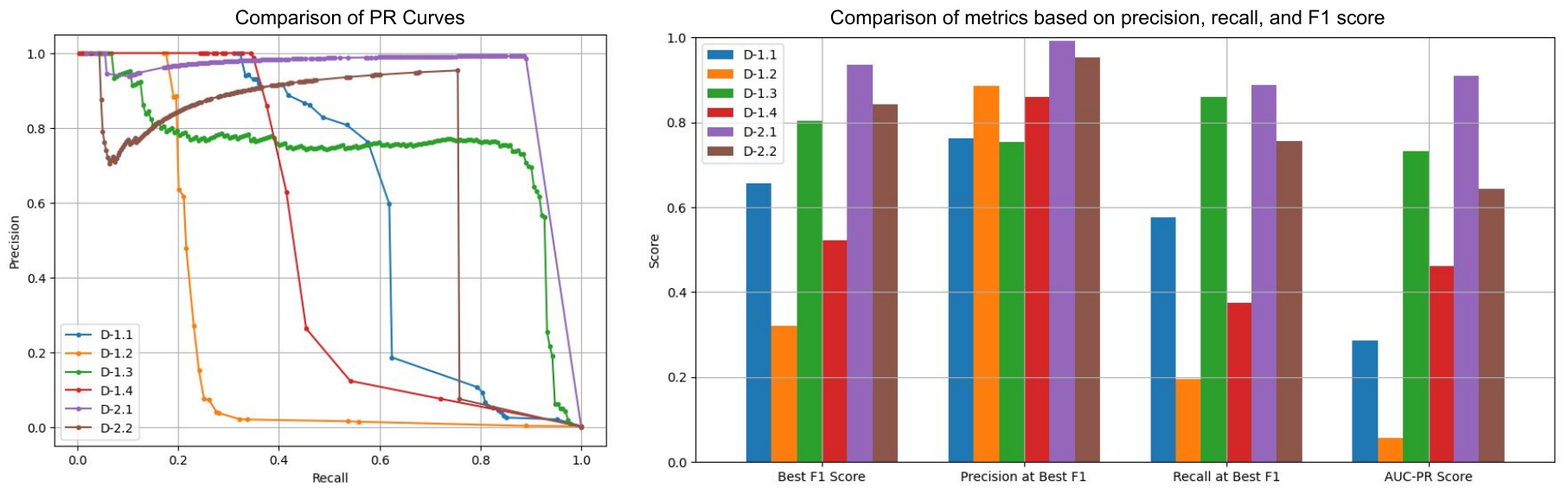}
    \caption{Anomaly Detection Performance of ReeFRAME.}
    \label{fig:pr_curves}
\end{figure*}

Typically, an anomalous agent would be instructed to travel to a designated 50m x 50m geographical area (or a set of such areas), stay there for a set period, and then return to their original path. ReeFRAME’s task was to detect these deviations from normal behavior and flag the corresponding agents as anomalous. Our team contributed by generating some of the anomalous agents in each dataset, following the anomaly generation instructions provided to our team, similar to the way they were provided to other teams.

To better mimic real-world scenarios, the datasets also include Weakly Labeled Groups (WLGs). These labels help narrow the search for anomalies by indicating groups of agents that might be involved in unusual activities. In this proof-of-concept experiment, each WLG included a group of agents, with a small subset of agents actually engaging in anomalous activities. ReeFRAME used these WLGs, along with the anomalous agent trajectories simulated by our team, to extract features and accurately detect the agents that were anomalous.

\begin{table}[!b]
\begin{tabular}{|c|c|c|c|c|}
\hline
               & \textbf{\begin{tabular}[c]{@{}c@{}}Best \\ F1 Score\end{tabular}} & \textbf{\begin{tabular}[c]{@{}c@{}}Precision \\ at Best F1\end{tabular}} & \textbf{\begin{tabular}[c]{@{}c@{}}Recall \\ at Best F1\end{tabular}} & \textbf{\begin{tabular}[c]{@{}c@{}}AUC-PR \\ Score\end{tabular}} \\ \hline
\textbf{D-1.1} & 0.657                                                             & 0.762                                                                    & 0.577                                                                 & 0.287                                                            \\ \hline
\textbf{D-1.2} & 0.321                                                             & 0.886                                                                    & 0.196                                                                 & 0.057                                                            \\ \hline
\textbf{D-1.3} & 0.803                                                             & 0.753                                                                    & 0.859                                                                 & 0.733                                                            \\ \hline
\textbf{D-1.4} & 0.523                                                             & 0.86                                                                     & 0.376                                                                 & 0.462                                                            \\ \hline
\textbf{D-2.1} & 0.936                                                             & 0.992                                                                     & 0.886                                                                 & 0.910                                                            \\ \hline
\textbf{D-2.2} & 0.842                                                             & 0.953                                                                     & 0.755                                                                 & 0.644                                                            \\ \hline
\end{tabular}
\caption{Quantitative Results showing the anomaly detection performance of ReeFRAME.} 
\label{tab:quant_res}
\end{table}

\subsection{Feature Extraction, Hyperparameter Tuning, and Results}
\label{ssec:feat_engg}

Given the nature of Reeb graphs, they are implicitly well-suited for modeling trajectories. However, it is essential to perform thorough feature engineering to accurately describe the nodes and edges of these Reeb graphs, enabling efficient comparison between the training and testing Reeb graphs. While we do not have Reeb graph construction parameters like $\tau$ and $\epsilon$ learnable with respect to a loss function, we can assign meaningful features to the Reeb graph nodes. These features can then be used to train downstream models to detect deviations from patterns of normalcy. Since Reeb graphs capture significant events within a trajectory, effectively describing the nodes in a Reeb graph can encapsulate the essence of an agent’s movements. As detailed in Section~\ref{ssec:time_reebs_desc}, trajectory data can be characterized using three types of features: 1) Location-Based; 2) Kinematics-Based; 3) Semantics-Based.

For the four datasets we worked with, the optimal set of features for each Reeb graph node included:
\begin{enumerate}
    \item Maximum stop duration of the agent at a given Reeb node 
    \item Maximum velocity of the agent at the Reeb node 
    \item Distance from each test Reeb node to the nearest train Reeb node
    \item Modes of transportation used by the agent at a given node
\end{enumerate}

We also explored the use of semantic features, such as the list of Points of Interest (POIs) around a given latitude-longitude to describe the nature of the location. Our goal was to identify a feature descriptor that could effectively flag one agent (or as few agents as possible) per Weakly Labeled Group (WLG), to the greatest extent possible. Ultimately, the feature set listed above proved to be the most effective.

Since it is common for a population to exhibit both bottom-up and top-down patterns of normalcy, we captured bottom-up patterns by comparing training and testing Reeb graphs on an agent-by-agent basis. For top-down patterns, we utilized the incremental Reeb graphs discussed in Section~\ref{ssec:time_reebs_desc}. We constructed a large Multi-Agent Reeb Graph (MARG) for the entire training population and compared each test trajectory against this MARG. In this experiment, each MARG node was described using GPS coordinates, maximum velocity, and maximum stop duration. Test Reeb nodes that deviated significantly from MARG nodes in terms of stop duration and GPS information were flagged as anomalies. We arrived at these feature vectors through automated hit-and-trial experiments on various weighted feature vector combinations, aiming to maximize the area under the Precision-Recall Curve (AUC-PR) for the generated anomalies while minimizing the number of agents flagged per WLG. Quantitative results on D-1.* and D-2.* datasets are presented in Table~\ref{tab:quant_res}. Figure~\ref{fig:pr_curves} shows the Precision Recall curves along with bar charts of other anomaly detection metrics. 
We also report the run time taken by both D-1.* and D-2.* datasets in Section~\ref{ssec:aws}, highlighting scalability of ReeFRAME.

This feature engineering methodology can become computationally expensive as the population size and metadata increase. We are currently working on developing a trainable framework to address this challenge (further details in Section~\ref{sec:conclusion}).

\begin{figure*}[!htbp]
    \centering
    \includegraphics[width=2\columnwidth]{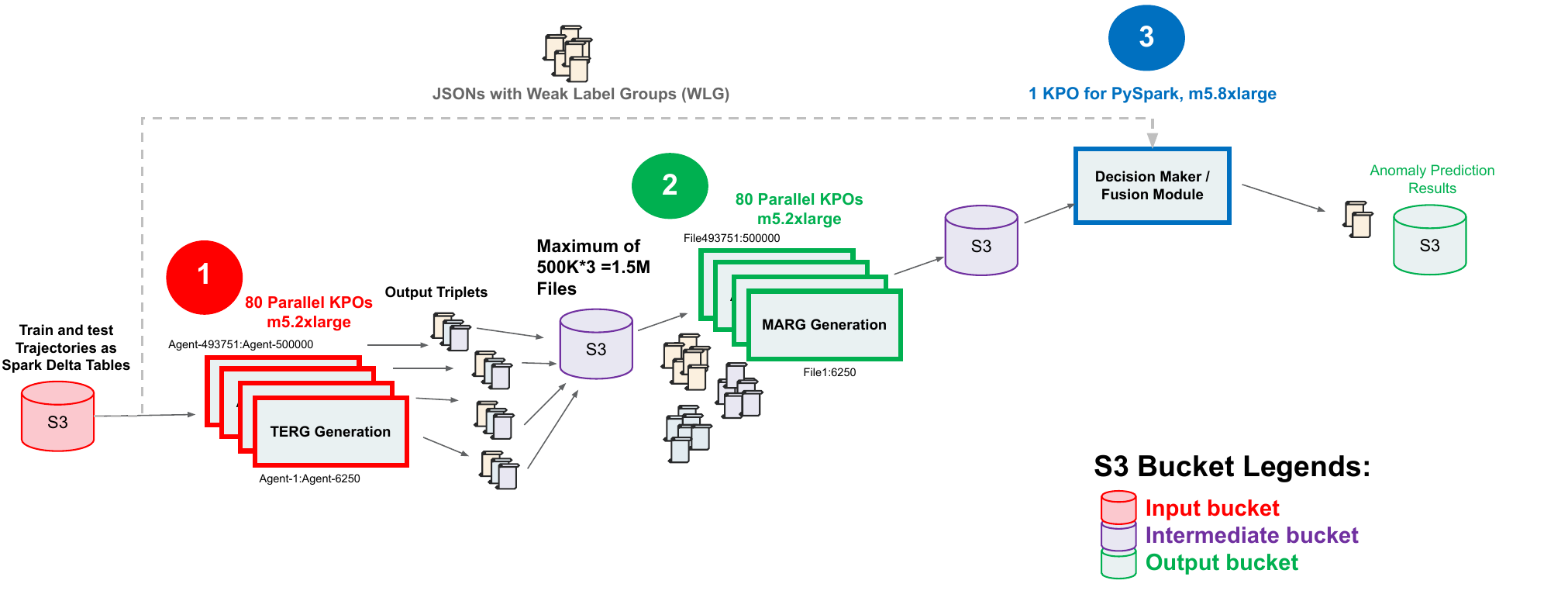}
    \caption{AWS Architecture Diagram, demonstrating the cloud readiness of ReeFRAME and the strategies that we used to run the pipeline on cloud.}
    \label{fig:aws_arch_diag}
\end{figure*}

\subsection{Scalable Inference on Cloud}
\label{ssec:aws}

The feature engineering and hyperparameter tuning for ReeFRAME were conducted using an on-premise CPU cluster equipped with 732 CPU cores and 1.25TB of RAM. However, the current inference pipeline has been optimized for cloud deployment and is packaged as an Airflow Directed Acyclic Graph (DAG). This DAG efficiently utilizes CPU nodes on an AWS EC2 cluster, balancing cost and performance. Figure~\ref{fig:aws_arch_diag} provides an overview of the AWS architecture used to process the datasets described in Section~\ref{ssec:datasets}. For the largest dataset, consisting of 500,000 agents, the DAG completed processing in under 12 hours.

In this section, we provide a detailed description of the compute and storage clusters employed to execute ReeFRAME in a cost-effective and time-efficient manner on AWS. While there are various ways to leverage cloud infrastructure, we describe the pipeline that we developed to process the large scale datasets that we developed in this paper. The train and test trajectories for each agent were provided as Delta tables partitioned by agent ID, accommodating up to 500,000 agents. Given the substantial size of the input and output data, Delta Lake's Delta tables were instrumental in managing these large datasets. All input data was organized as agent-partitioned Delta tables, and the final predictions are also packaged as a Delta table for consistency. 

One of the main challenges we faced was the overhead associated with reading the large Delta tables containing the train and test trajectories, as well as writing the output Delta tables with the predictions. The train trajectories dataset alone, sampled at 1Hz frequency, occupied up to 1TB of S3 storage. Transferring such a large dataset from S3 to an EC2 node was found to be a significant bottleneck. Procuring a high-capacity instance with 128 CPU cores to process all agents on one single node, for example, would have increased costs significantly due to the I/O overhead and compute overhead, adding another layer of complexity.

To tackle this challenge, we implemented a strategy where multiple low-capacity CPU nodes were provisioned, with each node responsible for processing a distinct subset of agents from the input Delta table. Since the input data was partitioned by agent, each agent’s trajectory was stored in a separate folder. We leveraged this structure to enable scalable data reading operations using PySpark's capability to manipulate Delta tables. By specifying the agent’s folder path in the \texttt{pandas.read\_parquet} function, we efficiently loaded the data for each agent. It is important to note that this approach requires a clean Delta table; the presence of unnecessary parquet files in the agent folders could necessitate an additional pre-processing step to clean the input Delta table. Fortunately, our input Delta table was already clean, eliminating the need for this preprocessing step. We deployed multiple parallel CPU-optimized, low memory compute instances, which are both cost-effective and readily available, to process subsets of agents concurrently.

In the MARG generation stage, all nodes in the agent-level TERGs with a stop duration greater than one hour were passed on. This stage involved using these stop points to create a pseudo MARG from the training TERGs. Each test TERG was then compared against the training MARG to generate agent-level and point-level predictions for every agent. The decision-maker, or fusion module, at the final stage, aggregated these predictions and saved the results into the output S3 storage bucket (represented in green in Figure~\ref{fig:aws_arch_diag}).

\section{Future Work and Discussion}
\label{sec:conclusion}

While ReeFRAME has demonstrated its effectiveness in modeling and detecting anomalies in large-scale human trajectory datasets, there are several avenues for future work that could further enhance its capabilities and applicability:

\begin{enumerate}
    \item \textbf{Differentiable Reeb Graph Parameters}: One of the main challenges in our current approach is the lack of differentiable parameters in the construction of Reeb graphs, such as $\tau$ and $\epsilon$. Developing a method to make these parameters differentiable with respect to a loss function would allow for more seamless integration with machine learning models, enabling end-to-end training and optimization.
    \item \textbf{Handling Noisy GPS Data}: GPS data often contains noise due to various factors, including signal loss, multipath effects, and device inaccuracies. Future work could focus on improving the robustness of ReeFRAME to noisy data, perhaps by incorporating advanced filtering techniques or leveraging machine learning methods to correct or smooth the noisy trajectories before Reeb graph construction.
    \item \textbf{Coordinated Movements}: Another area of interest is the detection and analysis of coordinated movements among groups of agents. Current methods focus primarily on individual trajectories, but many real-world scenarios involve agents moving in groups with shared objectives. Enhancing ReeFRAME to model and detect such coordinated behaviors could provide deeper insights into group dynamics and collective anomalies.
    \item \subchange{ \textbf{Weekday vs. Weekend Trajectory Patterns}: Although the current experiments utilize 'daily' trajectories without differentiating between weekdays and weekends, future work could explore this distinction. It is possible that treating weekday and weekend behaviors separately could improve TERG modeling, as agents may exhibit significantly different behaviors depending on the day. 
    }
    \item \textbf{Cloud-Based Enhancements}: While the current version of ReeFRAME is optimized for cloud deployment, there is potential for further improvements. For instance, exploring serverless architectures or containerized microservices could offer even greater scalability and flexibility. Additionally, implementing real-time anomaly detection capabilities on the cloud could enable ReeFRAME to be used in time-sensitive applications, such as traffic management or emergency response.
    \item \textbf{Merging Reeb Graphs}: Finally, there is the potential to explore different methods for merging Reeb graphs from different agents or time periods. This could be particularly useful for long-term monitoring of patterns or for analyzing the evolution of behavior over time. Developing efficient algorithms for graph merging while maintaining the integrity of the original structures could open up new possibilities for trajectory analysis.
\end{enumerate}

In summary, ReeFRAME is an efficient tool for analyzing human mobility patterns and detecting anomalies. We are working towards incorporating the above proposed enhancements, which could significantly broaden ReeFRAME's applicability and improve its performance in a variety of real-world scenarios.

\section{Acknowledgement}
This work is supported by the Intelligence Advanced Research Projects Activity (IARPA) via Department of Interior/ Interior Business Center (DOI/IBC) contract number 140D0423C0057. The U.S. Government is authorized to reproduce and distribute reprints for Governmental purposes notwithstanding any copyright annotation thereon. Disclaimer: The views and conclusions contained herein are those of the authors and should not be interpreted as necessarily representing the official policies or endorsements, either expressed or implied, of IARPA, DOI/IBC, or the U.S. Government.

\bibliographystyle{plain}
\bibliography{seek_tex/references}

\end{document}